\documentclass[runningheads]{llncs}
\def\compileIntroductionCameraReady{}

\ifdefined\compileIntroductionCameraReady
  \usepackage{eccv}
\else
  \usepackage[review,year=2026,ID=*****]{eccv}
\fi

\usepackage{eccvabbrv}  
\usepackage{graphicx}
\usepackage{float}
\usepackage{flafter}
\usepackage{placeins}
\usepackage{booktabs}

\usepackage[accsupp]{axessibility}  

\ifdefined\compileIntroductionCameraReady
  \usepackage{hyperref}
\else
  \usepackage[pagebackref,breaklinks,colorlinks,citecolor=eccvblue]{hyperref}
\fi

\usepackage{orcidlink}

\newcommand{\CosmosPredict}{Cosmos-Predict2.5~\cite{NVIDIA2025CosmosPredict}}

\newcommand{\CogVideoX}{CogVideoX~\cite{Yang2024CogVideoX}}

\begin{document}

\title{JEPA Guided Diffusion: Predictive Vision-Language Conditioning for Generative Traffic Forecasting} 

\titlerunning{JEPA Guided Diffusion for Generative Traffic Forecasting}

\author{Trinh Tra Giang Nguyen\textsuperscript{*} \and
Thanh Nguyen Vo\textsuperscript{*} \and
Nguyen Hoai Thuong Bui\textsuperscript{*} \and
Ha Duc Bui\textsuperscript{$\dagger$}}

\authorrunning{T. T. G. Nguyen et al.}

\institute{Ho Chi Minh City University of Technology and Engineering, \\Ho Chi Minh, Viet Nam \\
\email{nguyentrinhtragiang03@gmail.com} \email{nguyenvothanh04@gmail.com} 
\email{thuongbui7198@gmail.com} \email{ducbh@hcmute.edu.vn}}

\maketitle
\begingroup
\renewcommand{\thefootnote}{*}\footnotetext{First three authors contributed equally.}
\renewcommand{\thefootnote}{$\dagger$}\footnotetext{Corresponding author.}
\endgroup

\begingroup

\newcommand{\paperabstracttext}{%
Accurate traffic forecasting requires both understanding scene dynamics and synthesizing realistic future observations. Recent diffusion-based video generation models produce visually plausible predictions but require expensive end-to-end training and often entangle scene understanding with image synthesis. In this work, we propose a decoupled forecasting framework that separates future representation learning from video generation. A frozen V-JEPA encoder first extracts predictive latent representations from the observed traffic videos, capturing the underlying scene dynamics in a semantic latent space. A lightweight latent alignment module then projects these representations into the conditioning space of a frozen Cosmos diffusion module, enabling future video synthesis without retraining the large generative model. By freezing all foundation models and training only the lightweight alignment module, the proposed framework substantially reduces optimization complexity while preserving forecasting capability. Experimental results on the AI City Challenge 2026 Track 5 benchmark demonstrate that the proposed method achieved a score of 75.1297, ranking third in the competition. These results suggest that predictive world representations learned by V-JEPA can effectively guide downstream video generation, providing a practical and efficient alternative to end-to-end diffusion-based forecasting. Our code is available at \url{https://github.com/AlterraFa/JEPA-Guided-Diffusion}.%
}

  \begin{abstract}
    \paperabstracttext
  \end{abstract}

\endgroup

\begingroup

\section{Introduction}
\label{sec:intro}

Safe autonomous driving depends on accurately forecasting how the traffic environment may evolve. Such forecasting requires more than inferring the motion of individual road users, it must capture their uncertain behaviours, mutual interactions, and possible responses to the ego vehicle. By predicting these coupled future developments before acting, an autonomous vehicle can assess potential risks and select an appropriate course of action~\cite{Lefevre2014MotionPrediction}.

Traditionally, autonomous driving research addressed this prediction problem using physics based, manoeuvre based, and interaction aware motion models. These model families trade off prediction horizon, interaction awareness, uncertainty handling, and computational efficiency, limiting their ability to represent complex scene evolution. Generative world modelling has therefore emerged as a promising paradigm for capturing the multimodal and inherently unpredictable nature of real world traffic~\cite{Fu2024VideoWorldModels}. When instantiated as a video prediction model, it forecasts scene evolution densely at the pixel level, capturing not only the specified target agents but also their interactions with the surrounding environment. The generated rollout can be inspected directly, allowing errors in agent behaviour, spatial relationships, and physical consistency to be identified.

Building on this direction, a growing body of research on generative driving models has adopted diffusion and flow based video prediction, learning conditional distributions over plausible future traffic scenes from historical observations and control conditions~\cite{Wang2023DriveDreamer,Wang2024DriveWM,Lu2024WoVoGen,Gao2024Vista,Russell2025GAIA2}. However, the limitations of optimizing primarily for visual generation become particularly relevant to the \textit{Generative Traffic Forecasting} task in Track 5 of the AI City Challenge 2026, which uses the Woven Traffic Safety dataset~\cite{Kong2024WTS}. In this task, the provided descriptions specify only the expected pedestrian and vehicle behaviours, while the remaining scene evolution must be inferred from the visual history. Without sufficient traffic world understanding, optimizing for pixel level generative fidelity may produce visually realistic videos whose scene evolution remains behaviourally, logically, or physically inconsistent.

We therefore introduce a decoupled framework that separates traffic world understanding from pixel level video synthesis. Rather than relying on visual reconstruction to implicitly learn scene dynamics, we train a lightweight predictor that combines representations from a pretrained visual world model with descriptions to predict future latent representations. This separation enables the pipeline to optimize future traffic semantics and physical consistency directly, while video generation becomes an interpretable rendering of the predicted latent future for visualization and evaluation.

More specifically, our framework encodes historical frames using a pretrained V-JEPA encoder. A QFormer module, inspired by BLIP-2~\cite{Li2023BLIP2}, first fuses the resulting visual representations with the behaviour descriptions. A Llama module then refines the fused representation into a textual conditioning embedding, which conditions the video generator to produce the final rollout. This rollout is then refined using a lightweight postprocessing procedure before evaluation.

Our main contributions are as follows:
\begin{list}{\arabic{enumi}.}{%
  \usecounter{enumi}%
  \setlength{\topsep}{0pt}%
  \setlength{\partopsep}{0pt}%
  \setlength{\itemsep}{0pt}%
  \setlength{\parsep}{0pt}%
  \setlength{\leftmargin}{1.5em}%
  \setlength{\labelwidth}{1.2em}%
  \setlength{\labelsep}{0.3em}%
}
  \item We propose a decoupled framework that separates latent traffic world prediction from pixel level video generation, enabling future traffic evolution to be learned explicitly in latent space while retaining interpretable video outputs.
  \item We introduce a lightweight postprocessing procedure that complements the core architecture, improving the quality of the final predictions without requiring architectural modifications.
  \item We evaluate the complete system on Track 5 of the AI City Challenge 2026, where it achieves an overall score of \textbf{75.1297} and ranks \textbf{third} on the official leaderboard.
\end{list}

\endgroup

\begingroup

\section{Related Works}

\subsection{Joint Embedding Predictive Architectures}

Joint Embedding Predictive Architectures (JEPA) learn representations by predicting latent features rather than reconstructing observations in pixel space. I-JEPA~\cite{Assran2023IJEPA} introduced this principle for images by predicting the representations of masked target regions from visible context. V-JEPA~\cite{Bardes2024VJEPA} extended feature prediction to video and showed that a frozen encoder can support both appearance  and motion sensitive downstream tasks. These results motivate latent prediction as a means of learning semantic video features without requiring pixel reconstruction.

V-JEPA 2~\cite{Assran2025VJEPA2} scaled this approach to large video collections and demonstrated strong results in motion understanding, action anticipation, video question answering, and action conditioned planning. V-JEPA 2.1~\cite{MurLabadia2026VJEPA21} further introduced dense predictive supervision and deep self supervision to improve spatial and temporal grounding. Our work uses a frozen V-JEPA 2.1 encoder, but differs from prior JEPA applications in its downstream objective. Rather than applying the representation to recognition, planning, or direct latent prediction, we combine it with language to predict the conditioning context of a separate generative video model.

\subsection{Generative Traffic Forecasting and Multimodal Video Conditioning}

Generative traffic forecasting extends structured prediction from trajectories or occupancy to the evolution of the complete visual scene. General video models increasingly support semantic, visual, and multimodal conditioning. VideoPoet~\cite{Kondratyuk2023VideoPoet} represents text, images, video, and audio in a shared autoregressive model; diffusion transformer methods such as \CogVideoX{} improve text-video alignment; and \CosmosPredict{} supports text, image, and video conditioned world generation. However, these broad generation objectives do not specifically model the scene consistency and interaction requirements of traffic forecasting.

Driving world models introduce more specialized guidance. GAIA-1~\cite{Hu2023GAIA} models video, text, and actions jointly. DriveDreamer~\cite{Wang2023DriveDreamer} uses structured traffic information, and DriveDreamer-2~\cite{Zhao2025DriveDreamer2} adds language-model guidance for diverse driving video generation. DrivingWorld~\cite{Hu2024DrivingWorld} predicts spatial and temporal tokens for long horizon generation, while Epona~\cite{Zhang2025Epona} separates temporal dynamics from diffusion based rendering. Dynamics oriented methods such as MAD~\cite{Rahimi2026MAD} further incorporate historical motion features or separate motion prediction from appearance synthesis. However, connecting an independently pretrained JEPA encoder to a video generator presents a distinct challenge: JEPA produces predictive visual features, whereas the generator expects conditioning tokens learned in a different latent space under a different objective. Directly injecting these features therefore provides no guarantee that their predictive semantics can be interpreted by the generator. This incompatibility motivates a learned interface that combines JEPA representations with behavioural captions and maps them into the generator specific conditioning space without fine  tuning either backbone.

\endgroup

\begingroup

\section{Method}
\subsection{Problem Formulation}
\label{sec:problem-formulation}

For each traffic scene, let $H=\{h_t\}_{t=1}^{T_H}$ denote the observed history frames, $Q=(q_{\mathrm{ped}},q_{\mathrm{veh}})$ denote the provided scene captions and $F=\{f_t\}_{t=1}^{T_F}$ represents the ground truth future video sequence. Given the observed history $H$ and description $Q$, the objective is to generate a future video $\hat{F}=\{\hat{f}_t\}_{t=1}^{T_F}$ that accurately reflects the temporal progression of the traffic scene. Unlike conventional video generation task that mainly focus on producing visually realistic content, future traffic prediction requires preserving the underlying physical and temporal consistency of the scene, including agent trajectories, relative motion, and interaction pattern. Although the provided description $Q$ offer valuable semantic guidance about future events, they are inherently ambiguous and lack fine grained spatiotemporal details such as object trajectories, velocities, and interaction timing. A single caption can correspond to multiple future evolution

To address these challenges, we propose a predictive latent conditioning framework that augments semantic descriptions with spatiotemporal representations learned from the observed scene, enabling a pretrained video diffusion model to generate more consistent and physically plausible future traffic videos.

\subsection{System Overview}

\begin{figure}[!t]
  \centering
  \includegraphics[width=\linewidth]{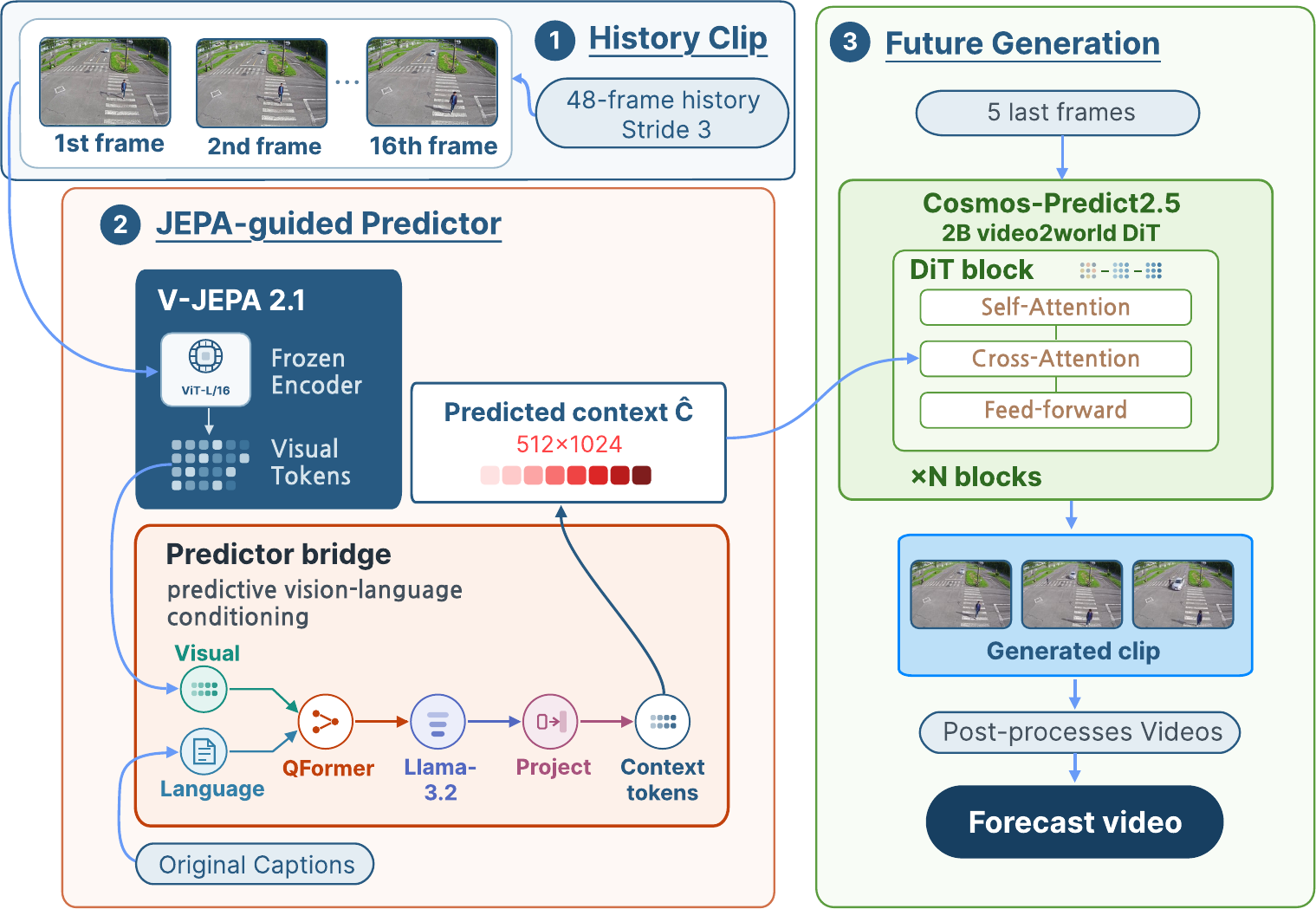}
  \caption{Inference overview. The trained predictor replaces Cosmos-Reason1 and directly supplies the cross attention context of the frozen Cosmos-Predict2.5 generator.}
  \label{fig:system-inference}
\end{figure}

Our framework integrates a predictive scene representation learned from a frozen world model with a pretrained video diffusion model for future traffic video generation, illustrated in Fig.~\ref{fig:system-inference}. Given the historical video sequence, we select the latest 48 frames of observations and temporally downsample the sequence by retaining one frame for every three frames, resulting in a compact 16 frame input clip. This sampling strategy reduces computational complexity while preserving long range temporal context of the traffic scene. The extracted clip is then processed by a frozen V-JEPA 2.1 encoder to obtain a predictive latent representation that captures the underlying spatiotemporal dynamics of the observed environment. The extracted latent is then fused with the provided caption and fed into a lightweight predictor, which aligns the fused visual-language representation with the conditioning space of the frozen Cosmos-Predict 2.5 DiT generator, effectively replacing the Cosmos-Reason1 conditioning component in the original pipeline. The aligned representation is subsequently combined with the last five historical frames and supplied to the generator for future frame generation. During training, the V-JEPA 2.1 encoder remains frozen, while Cosmos-Reason1 is used solely as a frozen teacher to provide target conditioning representations. Only the lightweight latent predictor is optimized to align the fused visual-language representation with this conditioning space, substantially reducing the training cost, as illustrated in Fig.~\ref{fig:system-training}. 

\begin{figure}[!t]
  \centering
  \includegraphics[width=\linewidth]{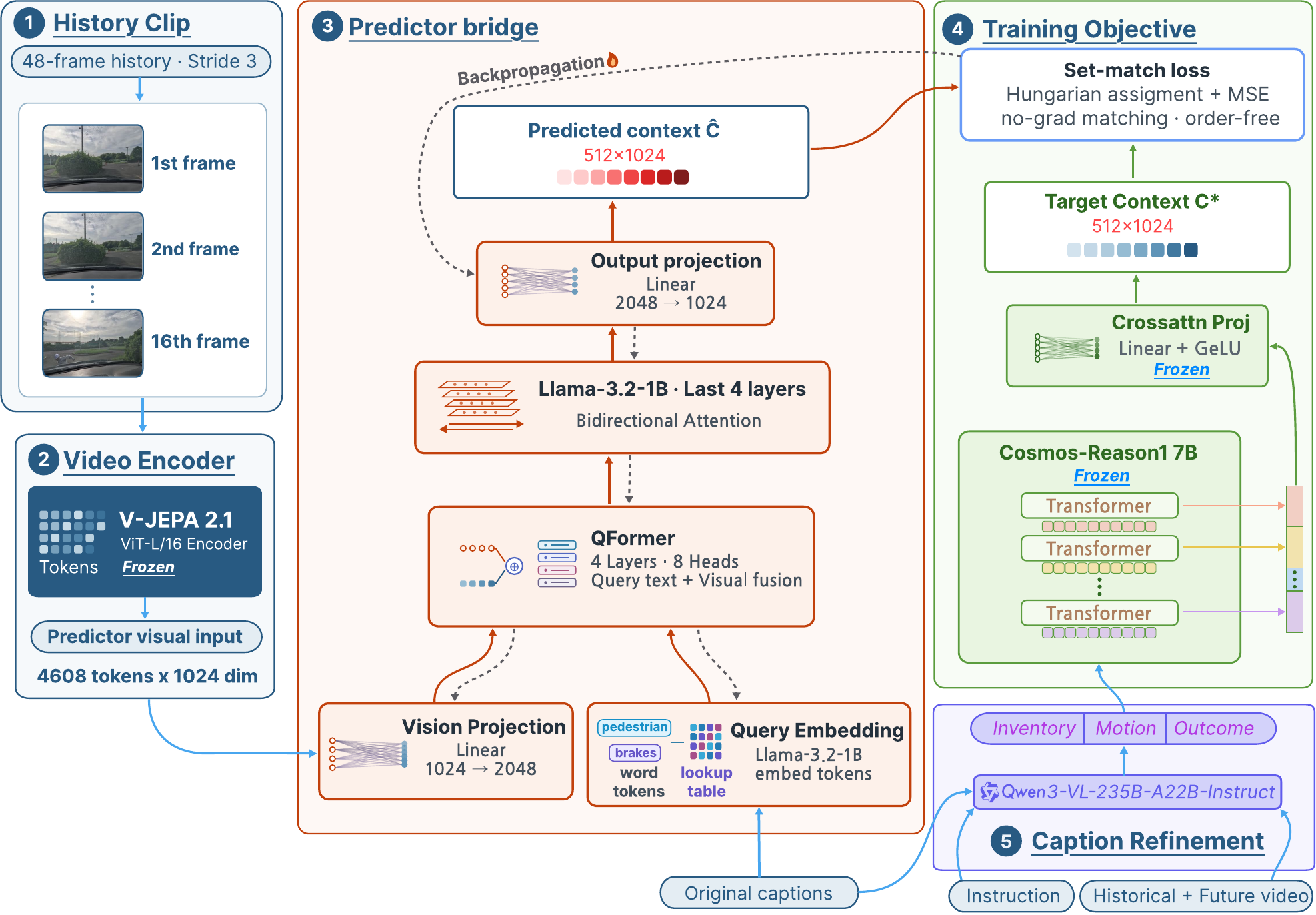}
  \caption{Training pipeline for learning the JEPA-guided conditioning context.}
  \label{fig:system-training}
\end{figure}

Finally, a postprocessing module is applied to enhance temporal consistency and visual quality of the generated videos, improving reconstruction metrics. Overall, the proposed framework decouples future understanding from video synthesis: V-JEPA provides predictive representations of scene dynamics, while Cosmos focuses on generating visually realistic future frames.

\subsection{Predictive Scene Representation with V-JEPA 2.1}

A frozen V-JEPA 2.1 ViT-L/16 encoder with 300M parameters represents the dynamics of the observed traffic scene. Its input consists of 16 frames resized to $384\times384$, giving a spatiotemporal input size of $16\times384\times384$. The encoder converts this clip into 4,608 spatiotemporal patch tokens, each with a feature dimension of 1,024:
\begin{equation}
Z_H=E_V(H_{VJEPA})\in \mathbb{R}^{4608\times1024}.
\end{equation}
The learned latent representation captures the implicit spatiotemporal patterns, including object motion, scene dynamics, and interaction cues. These properties make it suitable as complementary guidance for future traffic forecasting.

\subsection{Vision-Language Latent Alignment Predictor}

V-JEPA 2.1 latent is not directly compatible with the conditioning space of the Cosmos-Predict 2.5 DiT module. Hence, we introduce a lightweight multimodal latent predictor based on the last four layers of Llama-3.2-1B with a QFormer architecture~\cite{Li2023BLIP2}. The predictor takes both the visual representation and caption as inputs, enabling the alignment process to leverage complementary visual dynamics and semantic information.

\textbf{Vision and Language inputs.} A linear visual projection layer $W_v$ first maps the V-JEPA 2.1 features into the 2,048 dimensional hidden space of Llama-3.2-1B. Meanwhile, each caption is tokenized and padded to a fixed length of $L_Q=512$ token positions before being transformed by the Llama-3.2-1B input embedding layer:
\begin{equation}
Z=W_vZ_H\in\mathbb{R}^{4608\times2048},
\qquad
T_0=\operatorname{Embed}_{\mathrm{Llama}}(Q)
\in\mathbb{R}^{L_Q\times2048}.
\end{equation}

\textbf{Query conditioned fusion.} Directly feeding all visual and descriptions token to the language backbone would result in heavy computational burden. Therefore, we introduce a set of learnable query tokens $U_0\in\mathbb{R}^{512\times2048}$ as fixed output slots to aggregate multimodal information. Inspired by the QFormer in BLIP-2, these queries combine the caption with relevant visual evidence and produce a fixed length representation for future scene generation.

Let $S_0=[U_0;T_0]$ denote the initial joint sequence. The QFormer consists of four transformer layers. Each layer $l$ first applies shared self attention to the concatenated query and text representation:
\begin{equation}
\bar{S}_l=S_{l-1} + \operatorname{SelfAttn}_l(\operatorname{LN}_l^{(1)}(S_{l-1})).
\end{equation}
The resulting sequence is then separated into query and text streams:
\begin{equation}
(\bar{U}_l,\bar{T}_l)=\operatorname{Split}(\bar{S}_l).
\end{equation}
Only the query tokens interact with the visual representation through cross attention:
\begin{equation}
\tilde{U}_l=\bar{U}_l + \operatorname{CrossAttn}_l\!\left(
\operatorname{LN}_l^{(2)}(\bar{U}_l),
\operatorname{LN}_l^{(3)}(Z),
\operatorname{LN}_l^{(3)}(Z)
\right).
\end{equation}
The text tokens bypass visual cross attention and are preserved:
\begin{equation}
\tilde{S}_l=[\tilde{U}_l;\bar{T}_l].
\end{equation}
Finally, both streams are processed by a shared MLP:
\begin{equation}
S_l=\tilde{S}_l + \operatorname{MLP}_l(\operatorname{LN}_l^{(4)}(\tilde{S}_l)).
\end{equation}
Here, $\operatorname{LN}_l^{(1)}$--$\operatorname{LN}_l^{(4)}$ denote the layer-specific normalization modules applied before performing an operation.
After the final layer, RMS normalization is applied, and only the query tokens are extracted as the aligned multimodal representation
\begin{equation}
U_{QF}=\operatorname{Query}(\operatorname{RMSNorm(S_4)})\in\mathbb{R}^{512\times2048}.
\end{equation}

\textbf{Cosmos Conditioning Space Alignment.} The fused query tokens are further processed by the final four transformer layers of Llama-3.2-1B. Since the module is designed to learn a conditioning representation rather than generate autoregressive text, causal masking is disabled allowing each query token to attend bidirectionally to all other query tokens. A final linear layer $W_o$ projects the hidden dimension from 2,048 to 1,024:
\begin{equation}
\hat{C}=W_oB_{bi}(U_{QF})=\{\hat{c}_i\}_{i=1}^{512}\in\mathbb{R}^{512\times1024},
\end{equation}
where $B_{bi}$ denotes the bidirectional Llama layers. The resulting representation $\hat{C}$ matches the sequence length and embedding dimension of the post-projection context produced by Cosmos-Reason1, allowing it to condition the Cosmos-Predict 2.5 DiT directly.

\subsection{Predictor Training}

\begin{figure}[!t]
  \centering
  \includegraphics[width=\linewidth]{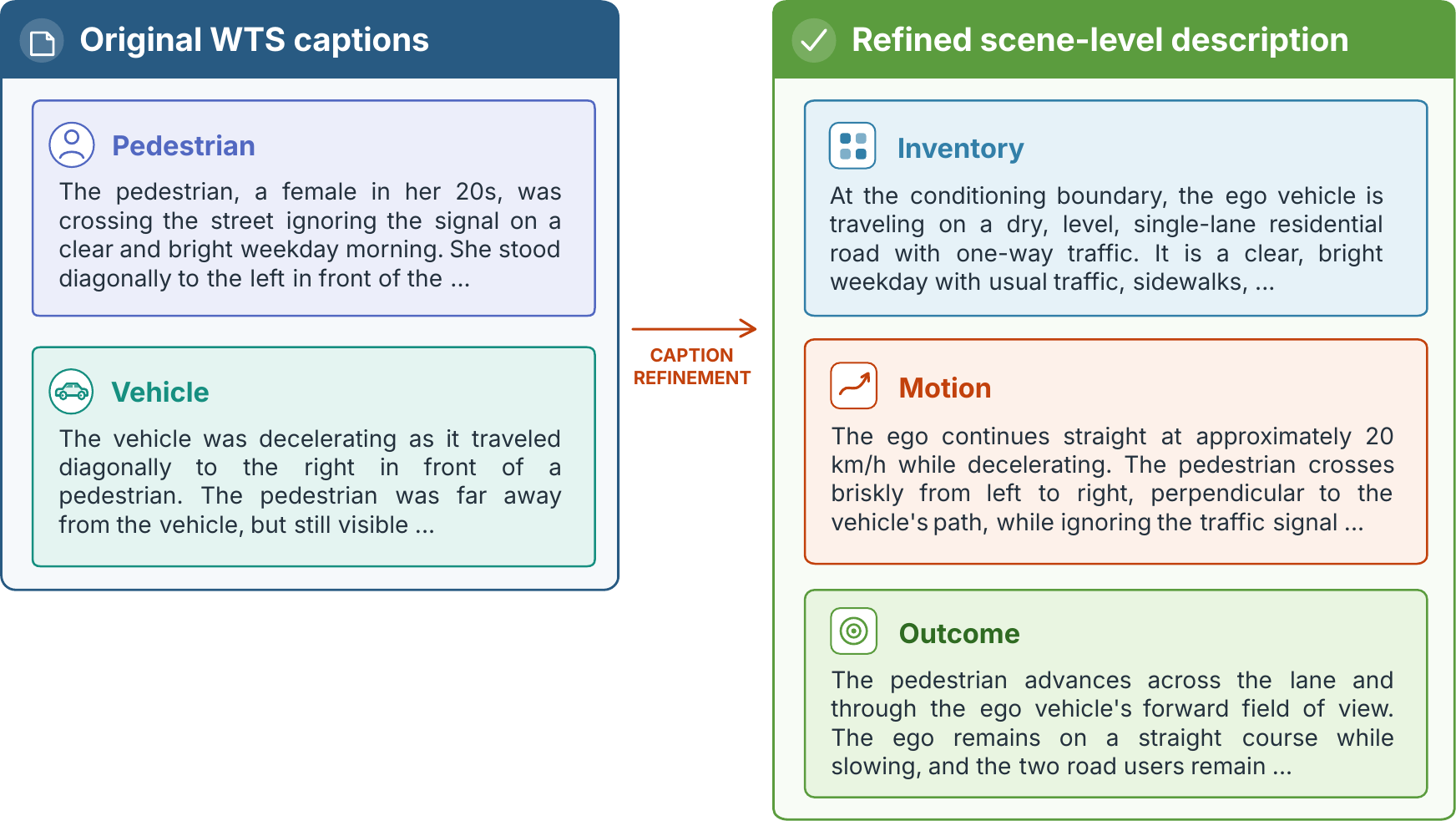}
  \caption{Example of the original and refined descriptions used for predictor training.}
  \label{fig:caption-refinement-example}
\end{figure}

\textbf{Caption Refinement Strategy.} To provide more physically grounded semantic guidance, we employ a VLM-based caption refinement pipeline. For each traffic scene in the training dataset, the input consists of the final 1 second of the historical video sequence, the complete ground-truth future clip, and the initial provided textual caption $Q$. These inputs are processed by Qwen3-VL-235B-A22B-Instruct~\cite{Bai2025Qwen3VL} with a structured prompting strategy to generate an enhanced future description containing three complementary aspects (an example shown in Fig.~\ref{fig:caption-refinement-example}):

\label{sec:caption-refine}
\begin{itemize}
  \item \textbf{Inventory}: Visible actors, objects, and their geometry at the conditioning boundary.
  \item \textbf{Motion}: Ego and actor motion throughout the target interval, including changes in relative distance.
  \item \textbf{Outcome}: Final actor positions, visibility, and contact state.
\end{itemize}

To satisfy the Cosmos-Reason1 prompt budget, Qwen further removes redundant wording from each refined description until it contains at most 300 tokens, while preserving the scene entities, motion, and outcome.

\textbf{Predictor Training.} For each training sample, the  frozen Cosmos-Reason1 encoder $E_R$ and cross attention projection $A$ transform the refined description $R$ into the target context representation:
\begin{equation}
C^*=A\!\left(E_R(R)\right)
=\{c_j^*\}_{j=1}^{N}\in\mathbb{R}^{N\times D},
\end{equation}
where $C^*$ represents the latent conditioning tokens extracted by Cosmos-Reason1 from the refined caption and serves as the supervision target.

The objective of training is therefore to align the output of the predictor $\hat{C}$ with the Cosmos context space, allowing the predictor to generate Cosmos compatible conditioning tokens directly from the observed video and caption.

Since Cosmos-Predict 2.5 consumes context tokens as unordered key-value representations in cross attention without positional encoding, the token ordering of $\hat{C}$ and $C^*$ is not fixed. Therefore, we perform permutation invariant matching between the predicted and target tokens using Hungarian assignment~\cite{Kuhn1955Hungarian}. Let $S_N$ denote the set of all permutations of the $N$ target token indices, where each permutation $\pi\in S_N$ defines a one-to-one assignment that pairs predicted token $\hat{c}_i$ with target token $c_{\pi(i)}^*$. The optimal assignment is

\begin{equation}
\pi^*
=\operatorname*{arg\,min}_{\pi\in S_N}
\sum_{i=1}^{N}
\left\lVert\hat{c}_i-c_{\pi(i)}^*\right\rVert_2^2.
\label{eq:hungarian-assignment}
\end{equation}

The Hungarian matching is computed without gradients, and the predictor is optimized using the matched mean squared error:
\begin{equation}
\mathcal{L}_{\mathrm{set}}
=\frac{1}{ND}\sum_{i=1}^{N}
\left\lVert\hat{c}_i-c^*_{\pi^*(i)}\right\rVert_2^2.
\label{eq:set-matching-loss}
\end{equation}

Even after convergence under the matching loss $\mathcal{L}_{\mathrm{set}}$, the predicted $\hat C$ and target $C^*$ remain different representations. While $C^*$ is generated from the refined caption alone, $\hat C$ incorporates both textual guidance and visual dynamics from the V-JEPA latent. Therefore, the predictor does not simply reproduce the text derived context, but learns a Cosmos compatible representation enriched with spatiotemporal information from the observed scene.

\subsection{Diffusion-Based Future Generation}


At inference, the frozen 2B Cosmos-Predict2.5 model receives the final five observed frames together with the predicted conditioning context $\hat C$. The frames initialize generation from the current scene, allowing the DiT to preserve its viewpoint and spatial structure. Starting from this visual state, $\hat C$ specifies how the scene is expected to evolve by encoding the predicted behaviours and interactions of traffic participants.

This guidance is introduced through cross attention in every DiT block. Given the video features $X_l$ at block $l$, the queries are derived from the evolving video representation, whereas the keys and values are derived from $\hat C$:

\begin{equation}
Q_l=X_lW_l^Q,
\qquad
K_l=\hat C W_l^K,
\qquad
V_l=\hat C W_l^V,
\end{equation}

\begin{equation}
\operatorname{CrossAttn}(X_l,\hat C)
=\operatorname{softmax}\!\left(\frac{Q_lK_l^{\mathsf T}}{\sqrt d}\right)V_l.
\end{equation}

Cross attention therefore conditions each update of the video representation on the predicted future while the observed frames retain the visual identity of the scene. The generated fixed length clip contains the five input frames followed by the synthesized future. We remove the input portion and truncate the remaining sequence to the required prediction length $T_F$, yielding the raw forecast $\hat F$.

\subsection{Postprocessing}
\label{sec:postprocess}

The forecasting task requires both plausible motion and consistency with the observed scene. In CCTV videos, most of the road layout and background remain stationary across the prediction interval, so the observed frames provide direct evidence for scene content that should persist. We use this property to restore static regions while retaining the foreground motion generated by Cosmos-Predict2.5. This assumption does not hold for dashcam video, where ego motion changes the background throughout the sequence. Applying the same restoration there could introduce ghosting, so dashcam forecasts are left unchanged.

Let $R=\{r_k\}_{k=1}^{K}$ denote the final $K=7$ observed frames, $\hat f_t$ a raw predicted frame, and $\tilde f_t$ its refined output. Because moving actors occupy each location only temporarily, the temporal median suppresses them while retaining the persistent background. We estimate this background image $B$ at each pixel $x$ as
\begin{equation}
B(x)=\operatorname*{median}_{1\leq k\leq K} r_k(x).
\end{equation}

Before restoring static detail from $B$, we first correct the appearance mismatch between the generated frame and the observed video. Specifically, we apply a whitening and coloring transform (WCT)~\cite{Li2017UniversalStyleTransfer} to align the color and brightness value of $\hat f_t$ with that of the final observation $r_K$. With transfer strength $\alpha=1.0$, the color-aligned frame is

\begin{equation}
c_t=\operatorname{WCT}_{1.0}(\hat f_t,r_K).
\end{equation}

The operator $\operatorname{Mask}_{25}$ converts an absolute RGB difference $D$ into a soft foreground mask:

\begin{equation}
\operatorname{Mask}_{25}(D)
=\operatorname{SmoothMorph}\!\left[
\operatorname{clip}_{[0,1]}\!\left(
\frac{\frac{1}{3}\sum_{c=1}^{3}D_c-12.5}{12.5}
\right)\right],
\end{equation}
where $\operatorname{SmoothMorph}$ removes isolated artifacts, fills small holes, and smooths mask boundaries through morphological closing and opening followed by Gaussian filtering. The mask is zero for mean RGB differences up to $12.5$, increases linearly between $12.5$ and $25$, and reaches one at $25$. Applying this operator to $|r_K-B|$ yields a reference foreground mask $M_R$ that identifies content present in the final observation but absent from the temporal-median background. We replace these regions in $r_K$ with the corresponding pixels from $B$ to obtain the actor-free reference $r_B$:

\begin{equation}
r_B=(1-M_R)\odot r_K+M_R\odot B.
\end{equation}

Directly replacing the generated frame with this reference would also overwrite its predicted structure and motion. Instead, a five-level Laplacian pyramid~\cite{Burt1983LaplacianPyramid} transfers only part of the observed high-frequency detail, such as edges and textures, while retaining the generated low frequency content. Here, $\mathcal{L}_{\ell}(I)$ denotes high-frequency level $\ell$ of image $I$, while $\mathcal{G}_{L}(I)$ denotes its coarsest low frequency level. With transfer strength $\beta=0.3$, the reconstructed image $q_t$ is defined by

\begin{equation}
\mathcal{L}_{\ell}(q_t)
=(1-\beta)\mathcal{L}_{\ell}(c_t)
+\beta\mathcal{L}_{\ell}(r_B),
\qquad
\mathcal{G}_{L}(q_t)=\mathcal{G}_{L}(c_t).
\end{equation}

Finally, $M_t=\operatorname{Mask}_{25}(|\hat f_t-B|)$ identifies foreground content in the prediction. Since reference detail could otherwise leak into moving actors, the intermediate frame $s_t$ attenuates this transfer with strength $\rho=0.5$. The median background is then restored only where $M_t$ indicates a static region:

\begin{equation}
s_t=(1-\rho M_t)\odot q_t+\rho M_t\odot c_t,
\qquad
\tilde f_t=M_t\odot s_t+(1-M_t)\odot B.
\end{equation}

Together, these operations refine complementary parts of the CCTV forecast: the Laplacian pyramid blend transfers observed high-frequency scene detail, the foreground mask limits this transfer within generated moving actors, and the temporal median background restores persistent static content. The resulting postprocessed frames therefore retain the motion and scene evolution predicted by the generator while remaining visually consistent with the observed environment.

\endgroup

\begingroup

\section{Experiment}

\subsection{Dataset}

We evaluate our method on the \textit{Generative Traffic Forecasting} task in the AI City Challenge 2026, which uses the Woven Traffic Safety dataset. The provided data are organized into external and internal subsets. The external subset contains 3,402 vehicle mounted videos, with 2,430 used for training and 972 for validation. The internal subset contains 249 synchronized traffic events captured from multiple viewpoints, resulting in 600 overhead camera videos and 210 vehicle view videos. Of these, 539 videos belong to the training split and 271 to validation. Internal videos are stored at $1920\times1080$ and approximately 30 fps, whereas external videos decode to $1280\times720$ and are predominantly recorded near 30 fps, with a smaller 60 fps subset.

Each event is annotated over five pedestrian--vehicle interaction phases. The external subset names them pre-recognition, recognition, judgement, action, and avoidance, while the internal subset uses indices 0 through 4 for the same progression. Every phase defines a start and end time and provides separate descriptions of pedestrian and vehicle behavior. These temporal annotations determine the observed history and target interval used to construct our forecasting samples.

\subsection{Implementation Detail}

Training the entire forecasting pipeline end-to-end on the WTS dataset is challenging due to the high computational cost. To address this limitation, our modular design enables the computationally intensive stages to be performed offline. Specifically, all training video segments are encoded into V-JEPA latent representations, while the caption-refinement strategy generates high-quality semantic descriptions from the raw captions. The refined descriptions are then encoded by Cosmos-Reason1 into conditioning representations that serve as training targets for the predictor. The resulting training samples, consisting of visual representations, raw captions, and refined semantic targets representations, are divided into training and validation sets. During training, only latent alignment predictor is updated. The predictor is trained using the proposed set-matching loss on a single NVIDIA GeForce RTX 5090, converging in approximately five hours with early stopping based on the validation loss. 

During inference, future video sequences are synthesized solely from history frames and raw captions on a single NVIDIA RTX6000. Full system configurations are detailed in Table~\ref{tab:training-generation-settings}.

\begin{table}[!t]
  \centering
  \caption{Training and generation settings.}
  \label{tab:training-generation-settings}
  \scriptsize
  \setlength{\tabcolsep}{3pt}
  \begin{subtable}[t]{0.49\linewidth}
    \centering
    \caption{Predictor training.}
    \label{tab:predictor-training-settings}
    \begin{tabular}{@{}p{0.41\linewidth}p{0.50\linewidth}@{}}
      \toprule
      Parameter & Configuration \\
      \midrule
      Optimizer & AdamW \\
      Batch size & 60 \\
      Base learning rate & $4\times10^{-5}$ \\
      QFormer learning rate & $8\times10^{-5}$ \\
      Weight decay & 0.04 \\
      Gradient clipping & 10 \\
      Scheduler & 5\% linear warm-up; cosine decay to $10^{-6}$ \\
      Maximum epochs & 50 \\
      Early-stopping patience & 5 epochs \\
      Seed & 42; deterministic execution \\
      \bottomrule
    \end{tabular}
  \end{subtable}
  \hfill
  \begin{subtable}[t]{0.47\linewidth}
    \centering
    \caption{Cosmos generation.}
    \label{tab:cosmos-generation-settings}
    \begin{tabular}{@{}p{0.43\linewidth}p{0.49\linewidth}@{}}
      \toprule
      Parameter & Configuration \\
      \midrule
      Checkpoint & Cosmos-Predict2.5 2B, post-trained video-to-world \\
      Conditioning frame rate & 30 fps \\
      Resolution & $1280\times720$ \\
      Diffusion steps & 35 \\
      Guidance scale & 7 \\
      Generation seed & 0 \\
      \bottomrule
    \end{tabular}
  \end{subtable}
\end{table}

\subsection{Evaluation Metrics}

The model performance is evaluated on the hidden Track 5 test set using six official metrics: PSNR and SSIM are used to assess pixel level reconstruction fidelity, LPIPS evaluates perceptual similarity, CLIP-S measures semantic consistency, and FID and FVD quantify the distributional similarity between generated and real samples at the frame and video levels, respectively. The final evaluation score is calculated as the unweighted average of the six normalized scores.

\subsection{Ablation Study}

To assess the contributions of the proposed decoupled architecture, we compare it with a VLM-based forecasting pipeline. In this baseline, the raw descriptions are refined into future oriented prompts using Qwen3-VL and then provided to the original Cosmos-Predict2.5 pipeline. Directly using the original captions as conditioning is suboptimal, as they provide limited guidance about future dynamics. Therefore, we adopt the caption refinement strategy described in Section~\ref{sec:caption-refine} to construct a stronger language based baseline. Since ground-truth future frames are unavailable during testing, Qwen3-VL-235B-A22B-Instruct is promted to utilize the 16 frame observed history, raw descriptions to produce a structured description containing scene composition, participant motion, and expected outcomes. The generated prompt is subsequently used as the conditioning input for Cosmos-Predict2.5 to generate the future video.

For a controlled comparison, both the proposed method and the VLM-based baseline receive the same observed history and original descriptions. Their final forecasts are generated with identical sampling settings and the same postprocessing procedure. They only differ in how the scene aware conditioning is constructed: the baseline uses VLM prompting, whereas our method directly predicts the conditioning context using V-JEPA2.1 features and raw captions.

Based on this controlled setup, we evaluate four variants obtained by combining the two conditioning strategies (VLM prompting and the JEPA-guided predictor) with and without the postprocessing. This comparison allows the effect of the proposed conditioning to be distinguished from the improvement introduced by postprocessing. The results are shown in Table~\ref{tab:method-comparison}.

\begin{table}[H]
  \centering
  \caption{Controlled comparison of conditioning and post-processing variants.}
  \label{tab:method-comparison}
  \scriptsize
  \setlength{\tabcolsep}{2pt}
  \resizebox{\linewidth}{!}{%
  \begin{tabular}{lcrrrrrrr}
    \toprule
    Method & Post-proc. & Final & PSNR $\uparrow$ & SSIM $\uparrow$ & LPIPS $\downarrow$ & CLIP-S $\uparrow$ & FID $\downarrow$ & FVD $\downarrow$ \\
    \midrule
    VLM-guided Cosmos & No & 72.0975 & 18.5038 & 0.5852 & 0.3152 & 0.9378 & 29.4280 & 25.1301 \\
    VLM-guided Cosmos & Yes & 73.4072 & 19.3147 & 0.6325 & 0.2894 & 0.9347 & 30.6414 & 24.9342 \\
    JEPA-guided Cosmos-Predict2.5 & No & 73.5120 & 19.1776 & 0.6306 & 0.2963 & 0.9356 & 28.6052 & 26.3175 \\
    \textbf{JEPA-guided Cosmos-Predict2.5} & \textbf{Yes} & \textbf{75.1297} & \textbf{19.7213} & \textbf{0.6504} & \textbf{0.2661} & \textbf{0.9452} & \textbf{26.5176} & \textbf{24.7875} \\
    \bottomrule
  \end{tabular}%
  }
\end{table}

\textit{Effect of JEPA-guided conditioning.}
Without postprocessing, JEPA-guided conditioning improves the final score by 1.4145 points, from 72.0975 to 73.5120, with higher PSNR and SSIM and lower LPIPS and FID, despite marginally lower CLIP-S and higher FVD. Since the generator and sampling settings are unchanged, this improvement demonstrates that the predictive latent representation learned by V-JEPA2.1 provides more effective scene dynamics guidance than language only conditioning. It further validates that future scene understanding can be decoupled from video synthesis through a learned latent conditioning pathway, allowing the generator to focus primarily on visual rendering while the latent representation captures future evolution.

\textit{Effect of postprocessing.}
Postprocessing improves the final score by 1.3097 points for VLM conditioning and by 1.6177 points for JEPA-guided conditioning. Although it slightly lowers CLIP-S and increases FID for the VLM baseline, it improves all six metrics for the JEPA-guided method. These results show that postprocessing procedure consistently enhances the generated forecasts for both conditioning strategies.

\textit{Combined effect of conditioning and postprocessing.}
With postprocessing, the JEPA-guided method reaches 75.1297, outperforming the VLM baseline by 1.7225 points and achieving the best value for all six metrics in Table~\ref{tab:method-comparison}. This consistent improvement confirms that the complete JEPA-guided pipeline provides the strongest overall performance among the evaluated variants.

\subsection{Official Leaderboard}

\begin{table}[H]
  \centering
  \caption{Official AI City Challenge 2026 Track 5 leaderboard.}
  \label{tab:official-leaderboard}
  \scriptsize
  \setlength{\tabcolsep}{2pt}
  \resizebox{\linewidth}{!}{%
  \begin{tabular}{rlrrrrrrr}
    \toprule
    Rank & Team & Final & PSNR $\uparrow$ & SSIM $\uparrow$ & LPIPS $\downarrow$ & CLIP-S $\uparrow$ & FID $\downarrow$ & FVD $\downarrow$ \\
    \midrule
    1 & Qyn & 76.4866 & 20.1202 & 0.6503 & 0.2463 & 0.9498 & 22.4089 & 21.7907 \\
    2 & SSUPER & 76.0385 & 19.7271 & 0.6300 & 0.2487 & 0.9384 & 21.1641 & 19.4627 \\
    \textbf{3} & \textbf{Latent Painter - UTE (Ours)} & \textbf{75.1297} & \textbf{19.7213} & \textbf{0.6504} & \textbf{0.2661} & \textbf{0.9452} & \textbf{26.5176} & \textbf{24.7875} \\
    4 & CHTTL\_A30 & 74.0544 & 18.8573 & 0.5970 & 0.2814 & 0.9423 & 23.7849 & 24.1105 \\
    5 & VGU\_ai\_lab & 73.2843 & 19.7360 & 0.6472 & 0.2942 & 0.9454 & 33.6081 & 29.3483 \\
    6 & Monash Connected Autonomous Vehicles & 72.0264 & 18.7460 & 0.6176 & 0.3074 & 0.9341 & 32.7695 & 31.8538 \\
    7 & SMART Lab & 71.4194 & 19.2749 & 0.6414 & 0.2793 & 0.9529 & 43.3759 & 38.9656 \\
    \bottomrule
  \end{tabular}%
  }
\end{table}

As shown in Table~\ref{tab:official-leaderboard}, our framework ranks third on the full Track 5 test set with a final score of 75.1297. It finishes 0.9088 points behind second place and 1.0753 points ahead of fourth. Its competitive performance across the six evaluation metrics shows that the final ranking reflects a balance of reconstruction fidelity, perceptual similarity, semantic agreement, and distributional quality.

\endgroup

\begingroup

\section{Conclusion}
We presented a decoupled traffic forecasting architecture that separates future scene understanding from video synthesis and performs effectively with a frozen generator. Its lightweight predictor combines abstract spatiotemporal knowledge from V-JEPA 2.1 with behavioural text and maps them into the conditioning embedding space of the Cosmos-Predict2.5 DiT. Lightweight post processing further enhances temporal consistency and visual quality. Together, these components achieved a score of 75.1297 and third place in Track~5 of the AI City Challenge 2026.
\endgroup

%
%
\bibliographystyle{splncs04unsrt}
\bibliography{main}
\end{document}